\documentclass[a4paper,twoside]{article}

\usepackage{epsfig}
\usepackage{subcaption}
\usepackage{calc}
\usepackage{amssymb}
\usepackage{amstext}
\usepackage{amsmath}
\usepackage{amsthm}
\usepackage{multicol}
\usepackage{pslatex}
\usepackage{apalike}
\usepackage{algorithm2e}
\usepackage[bottom]{footmisc}
\usepackage{booktabs}
\usepackage{array}
\usepackage{xcolor}
\usepackage{xurl}
\usepackage{tikz}
\usetikzlibrary{arrows.meta,positioning}
\usepackage{pgfplots}
\pgfplotsset{compat=1.18}
\usepackage[hidelinks]{hyperref}
\usepackage{SCITEPRESS}

\newcommand{\avff}{AVFF-R}
\hypersetup{
  pdftitle={DFD-Lab: A Modular Audio-Visual Deepfake Detection Pipeline},
  pdfauthor={Jan Rybarczyk, Mateusz Roszkowski, Jacek Komorowski},
  pdfsubject={Modular pipeline and audio-visual deepfake detection experiments},
  pdfkeywords={Deepfake Detection, Audio-Visual Learning}
}

\begin{document}
\title{DFD-Lab: A Modular Audio-Visual Deepfake Detection Pipeline}
\author{\authorname{Jan Rybarczyk, Mateusz Roszkowski and Jacek Komorowski}
\affiliation{Warsaw University of Technology}
\email{\href{mailto:jacek.komorowski@pw.edu.pl}{jacek.komorowski@pw.edu.pl}}
}
\keywords{Deepfake Detection, Audio-Visual Learning.}
\abstract{Comparing audio-visual deepfake detectors requires coordinating dataset adaptation, temporal input representation, model interfaces and experimental conditions. We present DFD-Lab, a modular pipeline that separates these responsibilities while supporting shared training and evaluation workflows. We integrate three implementations: Xception-based maximum-logit fusion, ResNet with temporal LSTM fusion, and our AVFF reimplementation. Experiments cover external testing, degradation-based training augmentation and evaluation-time corruption. On a filtered subset of Deepfake-Eval-2024, models trained on FakeAVCeleb attain baseline AUROC values of 0.504, 0.538 and 0.458. JPEG50 training augmentation raises these to 0.691, 0.605 and 0.570, respectively, while all three accuracies decrease. These results illustrate why training interventions, evaluation corruptions and metric-dependent outcomes should remain distinct within a common pipeline. The contribution is the integration of audio-visual processing, interchangeable detectors and configurable experimental workflows, supported by empirical case studies. The findings highlight the challenge of cross-dataset detection and the complementary information provided by ranking and classification metrics.}

\onecolumn \maketitle \normalsize \setcounter{footnote}{0} \vfill
\section{\uppercase{Introduction}}
\label{sec:introduction}

Audio-visual deepfake detection involves more than choosing a classification architecture. A comparison must also specify how faces are extracted, how audio is represented, how the two streams are associated in time, and how predictions are reduced to reported metrics. These responsibilities become particularly visible when integrating detectors that combine modalities at different stages. Frame-level fusion, recurrent temporal aggregation and learned cross-modal representations can share an experimental objective without sharing an internal computation.

DFD-Lab addresses this integration problem through a modular processing and evaluation pipeline. Dataset-specific adapters feed a common clip representation; detector implementations expose compatible inputs and classification outputs; and configuration separates data preparation from training and evaluation. The practical objective is to make these boundaries explicit, so that changing a detector does not require redefining the entire surrounding workflow. The contribution is a shared experimental infrastructure for detectors with different fusion and temporal processing strategies.

An equally important distinction concerns the experimental question being asked. Replacing the evaluation dataset tests transfer to another collection. Transforming training inputs changes the learning conditions. Applying a transformation only during evaluation probes sensitivity of an already trained model. These procedures can reuse preprocessing and metric code, yet their outcomes answer different questions. A common pipeline is useful only if it preserves that distinction instead of presenting all resulting numbers as interchangeable measures of robustness.

We present the DFD-Lab pipeline and evaluate three audio-visual detector implementations using complementary experimental protocols. The implementation is publicly accessible \cite{DFDLab2026}. Our analysis connects the processing and model interfaces to external testing, training interventions and evaluation-time transformations within one experimental workflow.

The paper makes three contributions. First, it describes a shared audio-visual processing path, including dataset adaptation, temporal association, clip storage and the representation passed to detectors. Second, it explains the detector and training boundaries using three heterogeneous implementations: Xception Max-Fusion, ResNet+LSTM and our AVFF reimplementation. Third, it presents experiments on external testing, degradation-based training augmentation and evaluation-time corruption, retaining both favorable and unfavorable metric changes.

In the external tests, JPEG50 training augmentation produces the highest AUROC among the examined conditions for each implementation, but reduces accuracy relative to its baseline. The source-domain corruption study complements this result by examining the same transformation families at evaluation time. Together, these experiments show how the role of a transformation and the choice of metric affect the interpretation of detector performance.

The scope is thus a modular system with experimental case studies, rather than a new detection architecture. The following sections position the work, describe the pipeline and protocols, present the results, and discuss their implications and limitations.

\section{\uppercase{Related Work}}
\label{sec:related}

Reusable evaluation frameworks organize the data processing, model execution and reporting needed to compare deepfake detectors. DeepfakeBench provides a unified benchmarking framework for deepfake detection \cite{Yan2023}. More directly related to audio-visual evaluation, DeepfakeBench-MM introduces an extensible modular codebase with standardized preprocessing, training and evaluation, integrating multiple multimodal datasets and detector implementations \cite{Zhao2025}. Its evaluation includes intra-dataset, cross-dataset and cross-pipeline protocols, as well as analyses of modality use and training augmentation \cite{Zhao2025}. These contributions establish shared infrastructure and multimodal evaluation as existing research directions. Within this setting, DFD-Lab focuses on the connection between a paired clip representation, interchangeable detector interfaces and three concrete experimental workflows. The present paper follows the processing path from dataset adaptation to metric reporting, then uses external testing, training augmentation and evaluation-time corruption to illustrate its operation. This focus connects the system description to specific experimental questions and to the interpretation of the resulting detector scores.

Audio-visual detection methods differ in how they combine evidence within and between modalities. Zhou and Lim introduce a joint detection model with separate audio and visual streams and a synchronization stream that connects their representations, addressing cases in which either or both modalities are manipulated \cite{Zhou2021}. This approach explicitly models the relationship between the streams rather than relying only on separate authenticity decisions \cite{Zhou2021}. AVFF instead learns audio-visual representations through a two-stage approach involving contrastive learning, autoencoding and complementary masking \cite{Oorloff2024}. These examples motivate distinguishing the representation shared by a pipeline from the fusion mechanism implemented by a detector. DFD-Lab accommodates this distinction by exposing paired inputs and binary logits at its model boundary while keeping fusion and temporal processing inside each implementation. Its case studies use Xception Max-Fusion, ResNet with long short-term memory (LSTM) processing, and AVFF-R. We use AVFF-R to denote our AVFF reimplementation, distinguishing it from the original authors' implementation and published results.

Dataset choice and input degradation provide complementary contexts for evaluation. FakeAVCeleb supplies multimodal audio-video deepfake data \cite{Khalid2021}, whereas Deepfake-Eval-2024 collects in-the-wild deepfakes circulated in 2024 \cite{Chandra2025}. In our experiments, FakeAVCeleb is the training collection and a filtered audio-visual subset of Deepfake-Eval-2024 supplies the external evaluation data. Perturbation-based evaluation has also been studied in visual face-forgery detection: DeeperForensics-1.0 incorporates diverse real-world perturbations, including compression and noise, to examine detector behavior under degraded inputs \cite{Jiang2020}. This provides context for treating input quality as an experimental variable alongside the choice of dataset. For audio-visual systems, our case studies consider visual degradation, audio degradation and temporal misalignment within the same workflow. We distinguish transformations applied to training inputs from those applied during evaluation of a fixed model. These interventions answer different questions from transferring a detector to another collection, even when they reuse preprocessing components. The experimental protocol therefore identifies both the transformation's role and the evaluation dataset.

Against this background, DFD-Lab contributes a focused system account supported by experimental case studies. Its shared processing and detector interfaces connect three fusion implementations to external evaluation and configurable transformation studies. Reporting both AUROC and accuracy then links the workflow to two complementary views of performance: score ranking and classification decisions. The contribution lies in this documented integration and empirical analysis, with particular attention to how intervention placement and metric choice shape the interpretation of results.

\section{\uppercase{DFD-Lab Pipeline}}
\label{sec:pipeline}

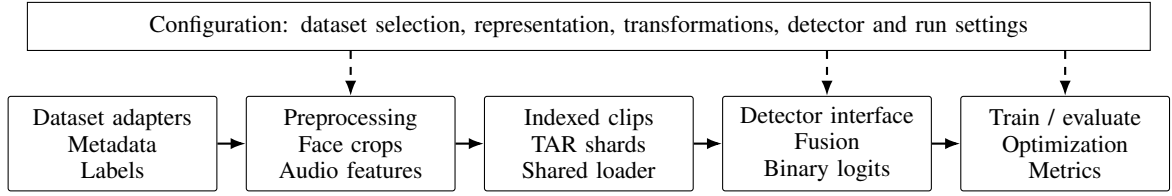
\begin{figure*}[t]
\centering
\begin{tikzpicture}[
 block/.style={draw,rounded corners=1pt,align=center,text width=2.55cm,minimum height=1.25cm,font=\small,inner sep=3pt},
 flow/.style={-{Latex[length=2mm]},thick},
 cfg/.style={draw,align=center,text width=14.5cm,font=\small,inner sep=5pt}
]
\node[block] (data) at (0,0) {Dataset adapters\\Metadata\\Labels};
\node[block] (prep) at (3.15,0) {Preprocessing\\Face crops\\Audio features};
\node[block] (store) at (6.3,0) {Indexed clips\\TAR shards\\Shared loader};
\node[block] (model) at (9.45,0) {Detector interface\\Fusion\\Binary logits};
\node[block] (run) at (12.6,0) {Train / evaluate\\Optimization\\Metrics};
\draw[flow] (data)--(prep);
\draw[flow] (prep)--(store);
\draw[flow] (store)--(model);
\draw[flow] (model)--(run);
\node[cfg] (config) at (6.3,1.55) {Configuration: dataset selection, representation, transformations, detector and run settings};
\draw[dashed,flow] (config.south -| prep.north)--(prep.north);
\draw[dashed,flow] (config.south -| model.north)--(model.north);
\draw[dashed,flow] (config.south -| run.north)--(run.north);
\end{tikzpicture}
\caption{DFD-Lab component boundaries and forward data flow. Configuration controls input preparation and run settings; the training layer additionally updates detector parameters.}
\label{fig:pipeline}
\end{figure*}

DFD-Lab separates dataset-specific preparation, shared representation, detector computation and experiment orchestration. Figure~\ref{fig:pipeline} summarizes these boundaries. The following description covers the audio-visual processing path and the shared interfaces used to organize experiments.

\subsection{Data Processing and Representation}

Dataset adapters isolate differences in metadata, file organization and labels before the shared processing path. The implementation contains adapters for FakeAVCeleb, Deepfake-Eval and pretraining data. Their purpose is to provide usable records to the preprocessor without making each detector interpret a dataset's original directory structure. This separation allows a dataset change to occur upstream of the model, although a new adapter must still make explicit decisions about label mapping and sample eligibility.

The experimental video-processing path uses MTCNN face detection, an additional crop margin of 0.2 and resizing to $299\times299$ pixels. Face cropping focuses the representation on facial appearance, providing a consistent spatial input for the visual branches of the detectors.

For audio, we use a sampling rate of 16~kHz and a 64-band mel representation, with an FFT size of 1024 and a hop length of 256. Audio features are associated with visual positions in a clip before being passed to a detector. This establishes a common temporal indexing convention, rather than requiring every model to decode a waveform independently. The representation is a design choice: architectures requiring raw waveforms or a different time-frequency resolution would need an additional conversion or a changed input contract.

Clips provide the unit of storage and model input. The preprocessing configuration uses windows of 25 frames with a stride of 20. Frames and audio features are serialized into TAR shards, using WebP quality 90 for images and half-precision audio arrays. The shared loader reconstructs tensors from these records. Offline preparation avoids repeating the entire extraction path in every training epoch, but commits subsequent runs to a particular preprocessing product. A change in crop generation, audio features or offline degradation therefore creates a different input artifact, not merely a different model setting.

Table~\ref{tab:interfaces} distinguishes the logical records from the tensors consumed by a detector. $B$ denotes batch size, $T$ the number of temporal positions, and $H$ and $W$ the relevant spatial or feature dimensions. Audio and visual dimensions need not be identical. The implementation's common forward interface returns two logits per input clip, leaving modality fusion and temporal aggregation inside the detector.

\begin{table*}[t]
\caption{Principal representation boundaries. Tensor shapes describe the common detector input after loading and layout conversion; temporal length and feature dimensions are configuration-dependent.}
\label{tab:interfaces}
\centering\small
\begin{tabular}{>{\raggedright\arraybackslash}p{2.5cm}>{\raggedright\arraybackslash}p{4.0cm}>{\raggedright\arraybackslash}p{7.9cm}}
\toprule
Boundary & Representation & Responsibility\\
\midrule
Adapter to processing & Source records and labels & Interpret dataset metadata and identify eligible audio-visual inputs.\\
Processing to storage & Indexed clips in TAR shards & Store associated image and audio content; preserve the connection to the prepared input product.\\
Loader to detector & Visual: $B\times T\times3\times H_v\times W_v$; audio: $B\times T\times1\times H_a\times W_a$ & Supply temporally indexed paired tensors; $T$ and feature dimensions depend on preparation and loading.\\
Detector to task & Logits: $B\times2$ & Encapsulate modality fusion and temporal reduction; retain a consistent real/fake class order.\\
Evaluation to analysis & Scores, classes, aggregate metrics & Compute ranking and classification metrics from detector outputs.\\
\bottomrule
\end{tabular}
\end{table*}

Preprocessing window length and the loader's requested sequence length are separate configuration parameters. The symbolic dimension $T$ in Table~\ref{tab:interfaces} expresses this separation between storage layout and detector input.

\subsection{Detector and Training Interfaces}

The common detector interface accepts the visual and audio tensors and produces binary classification logits. It also defines prediction, confidence and modality-feature access methods. This separates model-specific feature extraction from the shared training task while allowing each detector to implement its own fusion mechanism.

The training layer constructs the selected detector from configuration, passes batches through it, computes a configured loss, and manages optimization and metric reporting. Model module, class and parameters are separated from the surrounding training settings. This arrangement supports changing the detector without duplicating the whole training loop, while retaining model-specific initialization and pretraining.

The three implementations illustrate the boundary at different levels of fusion. Xception Max-Fusion has separate visual and audio branches based on the Xception architecture \cite{Chollet2017}. Let $z_{t,c}^{v}$ and $z_{t,c}^{a}$ be their logits for temporal position $t$ and class $c$. The implementation takes a maximum over modalities and then temporal positions, which can be written as
\begin{equation}
 z_c=\max_{1\leq t\leq T}\max\{z_{t,c}^{v},z_{t,c}^{a}\}.
 \label{eq:maxfusion}
\end{equation}
A softmax of the resulting class logits provides a clip score. Maxima for the real and fake classes can arise from different frames or modalities. The shared interface exposes the resulting logits without prescribing this fusion rule to other detectors.

ResNet+LSTM instead applies encoders from the residual-network family \cite{He2016} to the modality inputs and uses recurrent temporal processing before classification. Its internal state therefore depends on an ordered sequence rather than only on a maximum response. AVFF-R uses separate encoders and cross-modal mappings to combine original and mapped representations, with encoder pretraining preceding detector training. These distinctions are kept inside the detector and, where needed, its pretraining workflow; they do not require separate definitions of the external result table.

To integrate a further detector, its adapter must accept the paired tensors, return logits with the expected class order, and implement the interface methods used by the selected workflow. An architecture using another representation can add a conversion at the input boundary. For example, switching from Xception Max-Fusion to ResNet+LSTM changes the configured detector and its internal temporal aggregation, while retaining the dataset adapter, prepared clips and metric computation.

\subsection{Experimental Configuration and Outputs}

Configuration separates preprocessing choices, detector parameters and training or evaluation settings. The available code includes configurations for preprocessing, training, evaluation and encoder pretraining. Data transformations can consequently be assigned to the input preparation used by a training run or to the input preparation used during evaluation. Their placement, rather than their name alone, determines the experimental intervention.

The implementation supports image resizing and compression, audio perturbations and temporal offsets. These operate at different stages: an image codec changes image content, waveform noise precedes audio-feature extraction, and a temporal offset changes the association between modalities. Materializing such variants offline makes the selected input product explicit, but reduces the flexibility of changing transformations dynamically during training. A stored transformation configuration is therefore part of an experiment's provenance, not an incidental implementation detail.

The evaluation component converts detector logits into scores and predicted classes, then computes aggregate metrics and classification summaries. This keeps metric computation separate from model-specific feature extraction and fusion.

The pipeline thus separates three kinds of change: selecting another evaluation collection, changing the input product used for training, and changing evaluation inputs while keeping a trained detector fixed. This separation is the organizing principle for the case studies. It allows a transformation such as JPEG compression to be reused in distinct experiments without conflating its role in learning with its role in evaluating a fixed detector.

\section{\uppercase{Experimental Setup}}
\label{sec:setup}

We evaluate three detector implementations using external testing, degradation-based training augmentation and evaluation-time corruption. The experiments share the audio-visual processing and detector interfaces described in Section~\ref{sec:pipeline}.

\subsection{Datasets and Implementations}

We train the detectors on FakeAVCeleb \cite{Khalid2021} for binary real/fake classification with audio and visual inputs. External testing uses a filtered video subset of Deepfake-Eval-2024 \cite{Chandra2025}, comprising approximately 1,000 videos selected for compatibility with the audio-visual setup. This external collection provides a test of transfer beyond the source dataset.

The three detector labels refer to implementations within DFD-Lab. Xception Max-Fusion uses Xception-based modality branches and the aggregation in Equation~\ref{eq:maxfusion}. ResNet+LSTM combines residual features with temporal recurrent processing. AVFF-R is our reimplementation of AVFF \cite{Oorloff2024}, using LRS2 for pretraining \cite{Afouras2018}. The implementations represent late score fusion, recurrent temporal fusion and cross-modal feature fusion, respectively.

\subsection{Protocols and Transformations}

Table~\ref{tab:protocols} separates the three workflows. External baseline testing evaluates source-trained models on the filtered external subset. Training-augmentation experiments change the source training inputs and evaluate the resulting models externally. The corruption study instead changes source-domain evaluation inputs for models trained on clean inputs. ``Clean'' means the unaugmented reference condition; it does not imply lossless imagery or absence of the pipeline's ordinary preprocessing.

\begin{table*}[t]
\caption{Experimental workflows and the questions they address.}
\label{tab:protocols}
\centering\small
\begin{tabular}{>{\raggedright\arraybackslash}p{3cm}>{\raggedright\arraybackslash}p{5.5cm}>{\raggedright\arraybackslash}p{5.9cm}}
\toprule
Workflow & Intervention and evaluation & Evaluation question\\
\midrule
External testing & Source-trained baseline models evaluated on a filtered Deepfake-Eval-2024 subset & How do the trained detectors perform on another collection?\\
Training augmentation & Change source training inputs; evaluate the resulting models on the external subset & How does each training transformation affect external AUROC and accuracy?\\
Evaluation-time corruption & Change source-domain evaluation inputs for clean-trained models & How sensitive are fixed detectors to degraded audio-visual inputs?\\
\bottomrule
\end{tabular}
\end{table*}

We consider five degradation conditions. RES50 reduces spatial resolution by a factor of 0.5. JPEG50 applies JPEG compression with quality factor 50. AAC96K applies AAC audio compression at 96~kbit/s. SNR10 adds audio noise at a signal-to-noise ratio of 10~dB. DESYNC introduces an audio-visual temporal offset of up to 200~ms in either direction. These labels identify experimental interventions, not different datasets or separate detector architectures.

\subsection{Evaluation Metrics}

The area under the receiver operating characteristic curve (AUROC) and accuracy are reported on a zero-to-one scale. AUROC summarizes score ordering across decision thresholds; accuracy measures agreement of decisions with labels at the operating point used for a result. Reporting both distinguishes ranking performance from classification accuracy. For training augmentation, we also report absolute differences from the corresponding model's baseline.

\section{\uppercase{Experimental Case Studies}}
\label{sec:results}

The case studies connect the common pipeline to distinct evaluation questions. Table~\ref{tab:augmentation} presents the external baseline and all five training-augmentation conditions for each implementation. Table~\ref{tab:corruption} presents the source-domain corruption results.

\begin{table*}[t]
\caption{External-test results on our filtered Deepfake-Eval-2024 subset. Rows identify baseline or training-input conditions. \avff{} denotes our AVFF reimplementation. Best results in each column are shown in bold.}
\label{tab:augmentation}
\centering\small
\begin{tabular}{lrrrrrr}
\toprule
 & \multicolumn{2}{c}{Xception Max-Fusion} & \multicolumn{2}{c}{ResNet+LSTM} & \multicolumn{2}{c}{\avff{}}\\
Condition & AUROC & Accuracy & AUROC & Accuracy & AUROC & Accuracy\\
\midrule
Baseline & 0.504 & \textbf{0.655} & 0.538 & 0.601 & 0.458 & \textbf{0.651}\\
RES50    & 0.484 & 0.576 & 0.411 & 0.458 & 0.395 & 0.452\\
JPEG50   & \textbf{0.691} & 0.538 & \textbf{0.605} & 0.540 & \textbf{0.570} & 0.530\\
AAC96K   & 0.465 & 0.511 & 0.471 & \textbf{0.651} & 0.402 & 0.465\\
SNR10    & 0.543 & 0.577 & 0.512 & 0.493 & 0.390 & 0.494\\
DESYNC   & 0.496 & 0.562 & 0.453 & 0.540 & 0.382 & 0.482\\
\bottomrule
\end{tabular}
\end{table*}

\subsection{Cross-Dataset Model Comparison}

The baseline row in Table~\ref{tab:augmentation} illustrates the common external evaluation path for three different fusion implementations. AUROC is 0.504 for Xception Max-Fusion, 0.538 for ResNet+LSTM and 0.458 for AVFF-R. The corresponding accuracies are 0.655, 0.601 and 0.651. The AUROC values, close to or below 0.5, indicate weak ranking performance on the selected external data.

The metric ordering is itself instructive: ResNet+LSTM has the highest baseline AUROC but the lowest baseline accuracy among these implementations. The preferred model therefore depends on whether the evaluation prioritizes score ranking or classification decisions. This difference motivates retaining both metrics throughout the remaining experiments.

From the system perspective, this case study illustrates a dataset substitution around a stable detector interface. The external adapter changes the source of evaluation inputs while the paired representation and metric computation remain common to the three detector implementations.

\subsection{Degradation-Based Training Augmentation}

JPEG50 produces the highest external AUROC among the examined training conditions for all three implementations. Relative to baseline, Xception Max-Fusion increases from 0.504 to 0.691, ResNet+LSTM from 0.538 to 0.605, and AVFF-R from 0.458 to 0.570. The absolute changes are 0.187, 0.067 and 0.112, respectively.

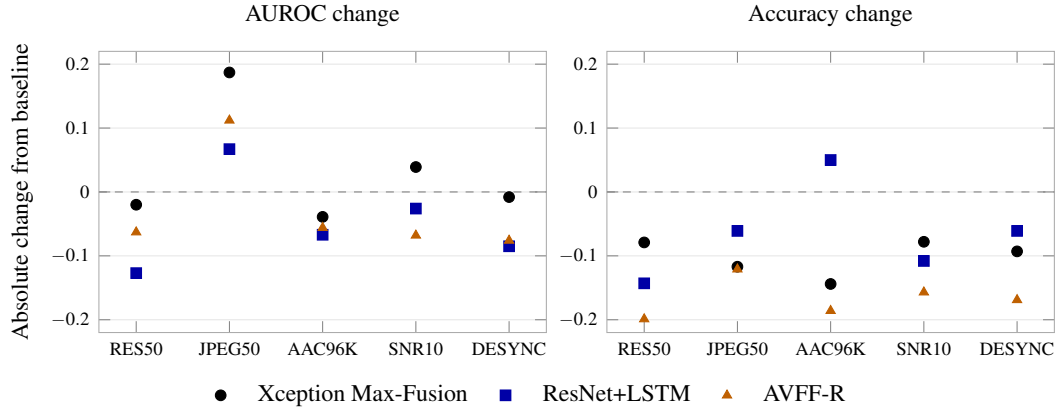
\begin{figure*}[t]
\centering
\begin{tikzpicture}
\begin{axis}[
 name=auc,width=7.5cm,height=5.3cm,
 title={AUROC change},ylabel={Absolute change from baseline},
 xmin=0.6,xmax=5.4,ymin=-0.22,ymax=0.22,
 xtick={1,2,3,4,5},xticklabels={RES50,JPEG50,AAC96K,SNR10,DESYNC},
 tick label style={font=\scriptsize},title style={font=\small},
 label style={font=\small},ymajorgrids,grid style={gray!20},
 axis line style={gray!65},legend style={draw=none,font=\small,column sep=10pt},
 legend columns=3,legend to name=auglegend]
\addplot[black,only marks,mark=*] coordinates {(1,-.020)(2,.187)(3,-.039)(4,.039)(5,-.008)};
\addlegendentry{Xception Max-Fusion}
\addplot[blue!65!black,only marks,mark=square*] coordinates {(1,-.127)(2,.067)(3,-.067)(4,-.026)(5,-.085)};
\addlegendentry{ResNet+LSTM}
\addplot[orange!75!black,only marks,mark=triangle*] coordinates {(1,-.063)(2,.112)(3,-.056)(4,-.068)(5,-.076)};
\addlegendentry{AVFF-R}
\addplot[gray,dashed,forget plot] coordinates {(.6,0)(5.4,0)};
\end{axis}
\begin{axis}[
 at={(auc.east)},anchor=west,xshift=0.8cm,width=7.5cm,height=5.3cm,
 title={Accuracy change},xmin=0.6,xmax=5.4,ymin=-0.22,ymax=0.22,
 xtick={1,2,3,4,5},xticklabels={RES50,JPEG50,AAC96K,SNR10,DESYNC},
 tick label style={font=\scriptsize},title style={font=\small},
 ymajorgrids,grid style={gray!20},axis line style={gray!65}]
\addplot[black,only marks,mark=*] coordinates {(1,-.079)(2,-.117)(3,-.144)(4,-.078)(5,-.093)};
\addplot[blue!65!black,only marks,mark=square*] coordinates {(1,-.143)(2,-.061)(3,.050)(4,-.108)(5,-.061)};
\addplot[orange!75!black,only marks,mark=triangle*] coordinates {(1,-.199)(2,-.121)(3,-.186)(4,-.157)(5,-.169)};
\addplot[gray,dashed] coordinates {(.6,0)(5.4,0)};
\end{axis}
\end{tikzpicture}
\par\smallskip
\ref{auglegend}
\caption{Changes relative to each model's external baseline, calculated from Table~\ref{tab:augmentation}. All five transformations are shown on identical vertical scales. JPEG50 increases AUROC but decreases accuracy for each implementation.}
\label{fig:augmentation}
\end{figure*}

All three JPEG50 accuracies move in the opposite direction. They decrease from 0.655 to 0.538, from 0.601 to 0.540, and from 0.651 to 0.530, respectively. Figure~\ref{fig:augmentation} places these changes beside the AUROC changes for every transformation, expressed as absolute differences from each model's baseline in Table~\ref{tab:augmentation}.

The remaining transformations do not provide a consistent improvement across implementations and metrics. RES50 lowers both reported metrics for all three models. SNR10 raises Xception's AUROC to 0.543 while reducing its accuracy to 0.577; its AUROC decreases for the other two implementations. AAC96K raises ResNet+LSTM accuracy to 0.651 while reducing its AUROC to 0.471. DESYNC decreases both metrics for all three implementations. Presenting the complete table avoids selecting only the favorable JPEG50 ranking results.

These findings distinguish improved score ranking from improved classification decisions. JPEG50 is favorable under the first criterion but unfavorable under the second. Describing it simply as improving detection would therefore obscure an important feature of the results.

The pipeline lesson is to record the location of an intervention and retain multiple evaluation measures. JPEG50 here changes training inputs; the following study applies it only during evaluation. A shared transformation label does not make these interventions equivalent.

\subsection{Evaluation-Time Corruption}
\label{sec:corruption}

Table~\ref{tab:corruption} illustrates the second placement of the same transformation families. We apply corruptions to FakeAVCeleb evaluation inputs for clean-trained models, examining how their outputs respond to changes in image quality, audio quality and temporal alignment.

\begin{table*}[t]
\caption{Evaluation-time corruption on FakeAVCeleb inputs using clean-trained models. Rows identify the transformation applied during evaluation; Clean is the unaugmented reference condition. Best results in each column are shown in bold.}
\label{tab:corruption}
\centering\small
\begin{tabular}{lrrrrrr}
\toprule
 & \multicolumn{2}{c}{Xception Max-Fusion} & \multicolumn{2}{c}{ResNet+LSTM} & \multicolumn{2}{c}{\avff{}}\\
Condition & AUROC & Accuracy & AUROC & Accuracy & AUROC & Accuracy\\
\midrule
Clean  & \textbf{0.990} & \textbf{0.952} & 0.967 & \textbf{0.992} & \textbf{0.652} & 0.723\\
RES50  & 0.712 & 0.557 & 0.961 & 0.828 & 0.640 & 0.714\\
JPEG50 & 0.554 & 0.663 & 0.968 & 0.904 & 0.643 & 0.722\\
AAC96K & 0.716 & 0.615 & \textbf{0.984} & 0.944 & 0.646 & 0.716\\
SNR10  & 0.643 & 0.524 & 0.773 & 0.686 & 0.350 & 0.671\\
DESYNC & 0.712 & 0.607 & 0.976 & 0.933 & 0.648 & \textbf{0.724}\\
\bottomrule
\end{tabular}
\end{table*}

For Xception Max-Fusion, AUROC changes from 0.990 in the clean row to 0.554 under JPEG50 and 0.643 under SNR10. ResNet+LSTM remains between 0.961 and 0.984 under RES50, JPEG50, AAC96K and DESYNC, while reaching 0.773 under SNR10. AVFF-R is between 0.640 and 0.648 for those four non-noise conditions and falls to 0.350 under SNR10, compared with 0.652 in the clean row. Audio noise reduces AUROC for all three implementations, whereas responses to the other transformations vary more strongly across detectors.

The two metrics again convey different information. For example, ResNet+LSTM's JPEG50 AUROC is 0.968 against a clean value of 0.967, whereas its accuracy changes from 0.992 to 0.904. For AVFF-R under SNR10, accuracy is 0.671 despite AUROC of 0.350. These cases reinforce the distinction between the quality of score ordering and the accuracy of classification decisions.

This case study demonstrates why transformations require both an intervention label and a dataset label. JPEG50 training augmentation improves external AUROC in Table~\ref{tab:augmentation}, whereas applying JPEG50 during source-domain evaluation substantially reduces Xception Max-Fusion's AUROC in Table~\ref{tab:corruption}. These experiments address different intervention--dataset combinations, so their outcomes should be interpreted within their respective protocols.

\section{\uppercase{Discussion and Limitations}}
\label{sec:discussion}

DFD-Lab separates a shared experimental workflow from model-specific computation. Dataset adapters isolate collection-specific organization, paired clips define the common input, and the detector interface encapsulates fusion and temporal processing. The integration of late score fusion, recurrent fusion and cross-modal feature fusion illustrates how heterogeneous detectors can share the surrounding data and evaluation workflow. This separation is particularly useful when an experimental question concerns the data or an input transformation rather than a change to the detector itself.

The results show that transformation placement is central to experimental interpretation. Training augmentation changes the conditions under which a model is learned, whereas evaluation-time corruption changes the inputs presented to a fixed model. External testing additionally changes the collection from which those inputs are drawn. DFD-Lab makes these choices explicit through separate data preparation and experiment configuration, allowing the same transformation family to support different evaluation questions.

The most consistent external-test observation is the combination of higher AUROC and lower accuracy after JPEG50 training augmentation. This is not contradictory: ranking quality and decisions at an operating point summarize different aspects of detector behavior. A practical consequence is that selecting an augmentation solely by AUROC can favor a configuration whose reported classification accuracy is lower. The complete metric pair is therefore more informative than a single claim of improved robustness.

The corruption study provides a complementary view of input sensitivity. SNR10 lowers AUROC for every implementation, while the visual and compression conditions produce more varied responses. This pattern motivates considering audio degradation alongside image degradation when evaluating audio-visual detectors. The variation across models also suggests that transformation settings should remain configurable rather than being treated as one fixed definition of robustness.

Two qualifications define the scope of these findings. First, external evaluation concerns a selected audio-visual subset of Deepfake-Eval-2024 rather than the full benchmark. Second, the comparisons concern the three implemented systems, including their respective pretraining and fusion choices; they should not be read as isolating the effect of architecture alone. Within this scope, the case studies demonstrate how a modular pipeline supports complementary experiments and exposes metric-dependent outcomes.

\section{\uppercase{Conclusion}}
\label{sec:conclusion}

We presented DFD-Lab as a modular audio-visual deepfake detection pipeline, organized around dataset adaptation, paired clip representation, detector interfaces and distinct training and evaluation workflows. Xception Max-Fusion, ResNet+LSTM and AVFF-R illustrate different internal computations behind the shared boundary. Their integration provides a common setting for studying data transformations and cross-dataset performance.

Our experiments cover external testing, degradation-based training augmentation and evaluation-time corruption. JPEG50 produces the highest external AUROC among the examined training conditions for all three implementations, while their accuracies decrease. Evaluation-time audio noise lowers AUROC for all three detectors. Together, these findings motivate preserving both metric distinctions and the role of each intervention in experimental reporting.

The implementation is publicly accessible in the DFD-Lab repository \cite{DFDLab2026}. Its modular organization provides a basis for extending the same experimental workflows to further audio-visual detectors and datasets.

\section*{\uppercase{Acknowledgements}}
\noindent
We gratefully acknowledge Polish high-performance computing infrastructure PLGrid (HPC Center: ACK Cyfronet AGH) for providing computer facilities and support within computational grant no. PLG/2025/017972

\bibliographystyle{apalike}
{\small\bibliography{references}}
\end{document}